\documentclass[11pt]{article}
\usepackage{graphicx} 
\graphicspath{{fig/}} 
\usepackage{acl}

\usepackage[most]{tcolorbox}
\usepackage{enumitem}
\usepackage{amsmath,amssymb,amsfonts}
\usepackage{enumitem}
\usepackage{xcolor}
\usepackage{tikz}
\usepackage{algorithm}
\usepackage{algpseudocode}
\usepackage{xcolor}
\usepackage{pifont}
\usepackage{microtype}

\definecolor{darkgray}{RGB}{60, 60, 60}           
\definecolor{lightgraybg}{RGB}{245, 245, 245}     
\definecolor{textgray}{RGB}{40, 40, 40}           

\definecolor{darkred}{RGB}{178, 34, 34}           
\definecolor{lightredbg}{RGB}{254, 242, 242}      
\definecolor{textgray}{RGB}{40, 40, 40}            

\definecolor{darkgreen}{RGB}{46, 125, 50}         
\definecolor{lightgreenbg}{RGB}{240, 253, 244}    
\definecolor{textgray}{RGB}{40, 40, 40}          

\definecolor{myblue}{RGB}{30, 144, 255}

\newtcolorbox{examplebox}[1]{%
    breakable,
    enhanced,
    colback=lightgraybg,
    colframe=darkgray,
    colbacktitle=darkgray,
    coltitle=white,
    title={#1},
    fonttitle=\bfseries\small,
    fontupper=\footnotesize\color{textgray},
    boxrule=0.8pt,
    arc=5pt,
    left=5pt,
    right=5pt,
    top=4pt,
    bottom=4pt,
    toptitle=3pt,
    bottomtitle=3pt,
    before skip=5pt,
    after skip=5pt
}

\newtcolorbox{redbox}[1]{%
    breakable,
    enhanced,
    colback=lightredbg,
    colframe=darkred,
    colbacktitle=darkred,
    coltitle=white,
    title={#1},
    fonttitle=\bfseries\small,
    fontupper=\footnotesize\color{textgray},
    boxrule=0.8pt,
    arc=5pt,
    left=5pt,
    right=5pt,
    top=4pt,
    bottom=4pt,
    toptitle=3pt,
    bottomtitle=3pt,
    before skip=5pt,
    after skip=5pt
}

\newtcolorbox{greenbox}[1]{%
    breakable,
    enhanced,
    colback=lightgreenbg,
    colframe=darkgreen,
    colbacktitle=darkgreen,
    coltitle=white,
    title={#1},
    fonttitle=\bfseries\small,
    fontupper=\footnotesize\color{textgray},
    boxrule=0.8pt,
    arc=5pt,
    left=5pt,
    right=5pt,
    top=4pt,
    bottom=4pt,
    toptitle=3pt,
    bottomtitle=3pt,
    before skip=5pt,
    after skip=5pt
}

\usepackage{times}
\usepackage{latexsym}
\usepackage{booktabs}
\usepackage{multirow}
\usepackage[table]{xcolor}
\usepackage{tabularx}
\usepackage{amsmath}

\usepackage[T1]{fontenc}

\usepackage[utf8]{inputenc}

\usepackage{microtype}

\usepackage{inconsolata}

\usepackage{graphicx}

\title{KnowGPO: Knowledge-Grouped Policy Optimization for Search Agents 
}

\author{
 \textbf{Siqi Chen\textsuperscript{1,2}},
 \textbf{Zhengxi Lu\textsuperscript{1}},
 \textbf{Yuchen Yan\textsuperscript{1}},
 \textbf{Xingyu Wu\textsuperscript{1,2}},
 \textbf{Qipeng Chen\textsuperscript{1}},
 \textbf{Xin Zhang\textsuperscript{2}},
\\
 \textbf{Aiting Liu\textsuperscript{2}},
 \textbf{Chao Deng\textsuperscript{2}},
 \textbf{Jie Liu\textsuperscript{2}},
 \textbf{Jin Ma\textsuperscript{2}},
 \textbf{Weiming Lu\textsuperscript{1}},
\\
 \textbf{Jun Xiao\textsuperscript{1}},
 \textbf{Yongliang Shen\textsuperscript{1}}\thanks{Corresponding author}
\\
 \textsuperscript{1}Zhejiang University,
 \textsuperscript{2}Tencent.
\\
  \texttt{\{siqichen,syl\}@zju.edu.cn, daniellwang@tencent.com}
}

\begin{document}
\maketitle
\begin{abstract}
Deep search agents solve knowledge-intensive tasks through long-horizon, multi-turn interactions, where assigning credit to individual search decisions is fundamentally difficult. Outcome-based reinforcement learning provides only a single terminal reward shared across an entire trajectory. Recent process-supervision methods aim to assign credit at finer granularity, but they typically anchor credit to either surface-level trajectory states or model-internal signals, both of which become unreliable in deep search where observations are heterogeneous and retrieved contexts are long and noisy. We propose \textbf{KnowGPO}, a reinforcement learning framework that aligns trajectories by the task-relevant knowledge they have acquired rather than by their surface observations. For each query, KnowGPO decomposes the information required to derive the answer into atomic knowledge conditions, and represents the state of a trajectory at each step by the subset of conditions it has satisfied. Trajectories sharing the same knowledge state are grouped for within-group comparison, and the incremental knowledge gain between consecutive steps serves as an environment-grounded process reward. Experiments on Xbench, BrowseComp-ZH, and GAIA show KnowGPO consistently outperforms outcome-only (GRPO) and process-supervision baselines (GiGPO, IGPO) by 3.8 to 4.2 points, with our 8B model rivaling or exceeding most 32B open-source systems. Our code is available at \url{https://github.com/ZJU-REAL/KnowGPO}.

\end{abstract}

\begin{figure*}[t] 
  \centering
  \includegraphics[width=\textwidth]{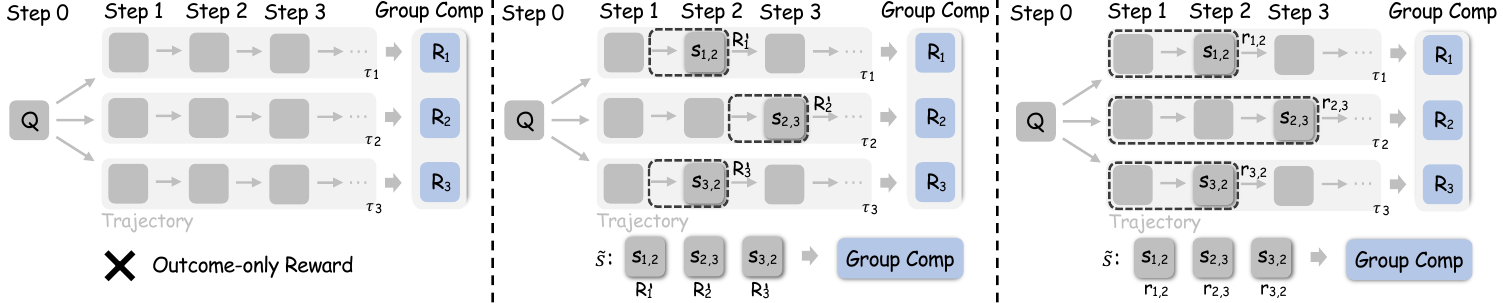} 
  \caption{\textbf{Comparison of credit assignment in deep search agents.} \textbf{Left:} GRPO uses episode-level advantages from terminal rewards ($R_1$, $R_2$, $R_3$) without intermediate feedback. \textbf{Middle:} GiGPO matches environment states ($\tilde{\mathbf{s}}$) but suffers from severe fragmentation under heterogeneous observations. \textbf{Right:} KnowGPO abstracts heterogeneous states into knowledge states ($\mathbf{s}_{i,t}$) and computes step-level advantages from incremental knowledge gains ($r_{i,t}$).}
  \label{fig:intro}
\end{figure*}
\section{Introduction}

LLM-based deep search agents have become a powerful paradigm for knowledge-intensive problem solving, coordinating multi-turn search, evidence inspection, query reformulation, and termination decisions through language-model reasoning~\citep{yao2022react,zhang2026landscapeagenticreinforcementlearning}.
Unlike conventional retrieval-augmented generation (RAG), which typically follows a one-pass retrieve-then-read pipeline~\citep{lewis2021retrievalaugmentedgenerationknowledgeintensivenlp,gao2024retrievalaugmentedgenerationlargelanguage,zhao2024retrievalaugmentedgenerationaigeneratedcontent}, deep search agents iteratively formulate queries, inspect evidence, update beliefs, and decide when to stop~\citep{nakano2021webgpt,li2025searcho1agenticsearchenhancedlarge}. This makes deep search a long-horizon sequential decision problem, where success depends on adaptive decisions about \textit{what to search, what evidence to keep, and when to terminate}. 
Because supervised fine-tuning mainly imitates surface behavior ~\citep{nakano2021webgpt,schick2023toolformerlanguagemodelsteach} and cannot optimize such strategic decisions, reinforcement learning has become central to training competitive deep search agents~\citep{Guo_2025,jin2025search,song2025r1}.

However, applying reinforcement learning to deep search agents encounters three coupled difficulties in credit assignment (Figure \ref{fig:intro}).
First, terminal outcome rewards used by group-relative methods such as GRPO~\citep{shao2024deepseekmathpushinglimitsmathematical,jin2025search,song2025r1,lu2025deepdive,li2025websailornavigatingsuperhumanreasoning,liu2025webexplorerexploreevolvetraining} are uniformly broadcast to every action, so a single final reward cannot reveal which intermediate decisions were useful, redundant, or harmful. 
Second, recent process-supervision methods such as GiGPO~\citep{feng2025groupingrouppolicyoptimizationllm} that align trajectories by their intermediate states require state recurrence across trajectories, yet deep search rollouts seldom revisit the same observable state even when they acquire the same evidence.
Third, dense process rewards derived from changes in the model's predicted answer probability, as in IGPO~\citep{wang2026informationgainbasedpolicyoptimization}, or from learned critics and rubric-based reward models~\citep{shi2025searchrefinethinkfacilitating,lee2025printsrewardmodelinglonghorizon,zhang2025criticsearchfinegrainedcreditassignment,zhang2026chainingevidencerobustreinforcement,lu2026sdar}, provide denser signals but are driven by model-internal beliefs or learned proxies rather than by externally verified knowledge.

Deep search trajectories vary widely at the level of queries, pages, and reasoning traces, but converge on the same underlying quantity: the task-relevant knowledge they have accumulated toward the correct answer. This knowledge, not observations or model confidence, provides the natural signal for credit assignment in long-horizon search.

We propose \textbf{Knowledge-Grouped Policy Optimization (KnowGPO)}, a RL framework that aligns trajectories by the task-relevant knowledge they have acquired rather than by their surface observations.
For each query, we decompose the information required to derive the correct answer into a set of atomic knowledge conditions. At every step of a trajectory, we represent its state by the subset of conditions it has satisfied, called its \textit{knowledge state}. We further define \textit{incremental knowledge gain}, the increase in satisfied conditions between consecutive steps, as a process reward that scores each action's immediate contribution. While knowledge-state grouping ensures only semantically comparable steps are contrasted, incremental gain provides the dense signal within each group, enabling fine-grained credit assignment without a learned critic. Trajectories sharing the same knowledge state are grouped for within-group comparison, yielding step-level advantages that combine with episode-level outcome advantages through a single mixing coefficient. The result is fine-grained credit assignment for long-horizon deep search without observation matching, without dependence on the model's own confidence, and without a learned reward model.


We train Qwen3-8B with KnowGPO on three challenging deep search benchmarks: Xbench, BrowseComp-ZH, and GAIA. KnowGPO improves over outcome-only GRPO by 2.0\% to 5.9\% and consistently outperforms recent process-supervision baselines including GiGPO and IGPO, with the resulting 8B agent matching or surpassing open-source systems of up to 32B parameters. 


In summary, our contributions are threefold:
\begin{itemize}
    \item We propose KnowGPO, a reinforcement learning framework that performs credit assignment in a query-specific knowledge state space rather than over surface observations, enabling within-group comparison across semantically equivalent trajectories.

    \item We introduce incremental knowledge gain as a process reward grounded in the environment, providing dense step-level supervision without observation matching, learned critics, or reliance on model-internal probabilities.

    \item We demonstrate that KnowGPO outperforms outcome-based and recent process-supervision baselines by 3.8\% to 4.2\% across three deep search benchmarks, with an 8B agent matching or surpassing open-source systems of up to 32B parameters.
\end{itemize}

\section{Related Work}


\subsection{Reinforcement Learning for Search Agent}
Reinforcement learning with verifiable outcome rewards has become the dominant paradigm for post-training deep search agents. Early systems train models to interleave search and reasoning through outcome-based optimization on multi-hop QA~\citep{jin2025search,chen2025svgenius,wang2025ragenunderstandingselfevolutionllm,zheng2025deepresearcherscalingdeepresearch, wu2025webdancerautonomousinformationseeking}, showing that group-relative methods such as GRPO scale to multi-turn tool use without learned value functions. Subsequent work scales this paradigm along two axes: harder data through knowledge graphs, web exploration, and synthetic multi-hop generation~\citep{lu2025deepdive,li2025websailornavigatingsuperhumanreasoning,liu2025webexplorerexploreevolvetraining,li2025websailorv2bridgingchasmproprietary,du2026openseekerdemocratizingfrontiersearch,gao2025turnsunlockinglonghorizonagentic,chu2026redsearcher,tongyideepresearchteam2026tongyideepresearchtechnicalreport, miromindteam2026mirothinkerpushingperformanceboundaries}, and refined reward shaping and policy gradient stability for long-horizon rollouts~\citep{wu2026deepsearchovercomebottleneckreinforcement,dong2025agenticreinforcedpolicyoptimization,luo2025infoflowreinforcingsearchagent, tang2025turnlimitstrainingdeep}. Despite these advances, the supervisory signal in most systems remains a binary reward at the end of each trajectory, leaving the contribution of search decisions inside long rollouts unidentified.

\subsection{Credit Assignment for Search Agents}
A growing line of work aims to address long-horizon credit assignment through process-level supervision. GiGPO~\citep{feng2025groupingrouppolicyoptimizationllm} extends GRPO with anchor-state grouping, computing step-level advantages by matching trajectories that revisit the same environment state; this mechanism requires reliable state recurrence, which is rare in deep search where retrieved contents and reasoning histories are open-ended. IGPO~\citep{wang2026informationgainbasedpolicyoptimization} instead derives dense step rewards from the model's own probability of producing the correct answer, sidestepping state matching but anchoring the signal to internal confidence that can be biased by long, noisy retrieved contexts. A parallel direction introduces externally grounded supervision through learned reward models or rubric decompositions~\citep{zhang2026chainingevidencerobustreinforcement,zhang2025criticsearchfinegrainedcreditassignment,lee2025printsrewardmodelinglonghorizon,shi2025searchrefinethinkfacilitating}, providing denser feedback than outcome rewards but relying on learned proxies whose calibration is itself a research question, and without producing step groupings that make semantically equivalent decisions directly comparable across trajectories. KnowGPO instead anchors credit assignment to an externally verifiable representation of progress, the set of satisfied atomic knowledge conditions, and uses this representation as both the grouping criterion for within-group comparison and the basis of a process reward grounded in the environment.

\section{Method}

\subsection{Problem Formulation}
\label{sec:problem}

We formulate deep search as a multi-turn sequential decision process. Given a query $Q$ and a reference answer $y^*$, an agent interacts with a retrieval environment over time steps $t \in \{1, \dots, T\}$. At each step the agent produces a reasoning trace and an action $a_t \in \{\texttt{search}, \texttt{answer}\}$ from its policy $\pi_\theta(\cdot \mid h_t, Q)$, where $h_t = (a_1, o_1, \dots, a_{t-1}, o_{t-1})$ is the interaction history and $o_t$ is the observation returned by the environment (e.g., retrieved documents for \texttt{search}, or a terminal signal for \texttt{answer}). A complete trajectory is denoted $\tau = (a_1, o_1, \dots, a_T, o_T)$ and terminates with a binary outcome reward $R(\tau) = \mathbb{I}(\text{ans}(\tau) = y^*)$. We build on group-relative policy optimization, which estimates advantages from a group of $N$ trajectories sampled for the same $Q$ and avoids learning a separate value network. The challenge addressed throughout this section is that $R(\tau)$ alone provides no signal for distinguishing useful, redundant, or harmful steps inside a long-horizon trajectory.

\begin{figure*}[t] 
  \centering
  \includegraphics[width=1.0\linewidth]{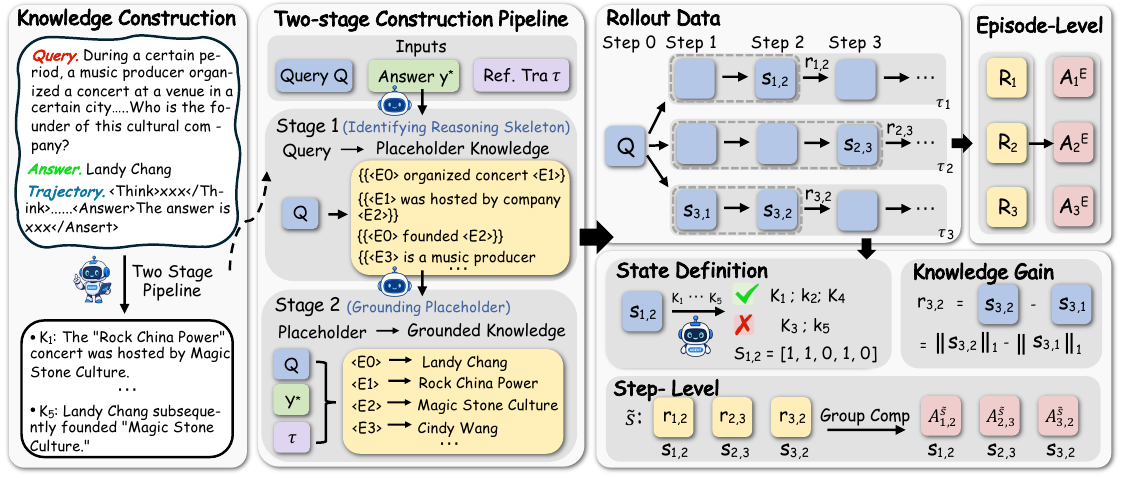} 
  \caption{\textbf{Overview of the KnowGPO framework.} \textbf{Left:} Knowledge condition construction via a two-stage pipeline, illustrated with an example. \textbf{Right:} Rollout trajectories are grouped by matched knowledge states; step-level advantages are computed from incremental knowledge gains via group-relative comparisons.}
  \label{fig:method}
\end{figure*}

\subsection{Knowledge State}
\label{sec:knowledge_state}


A central design choice in KnowGPO is to abstract the progress of a search trajectory through the task-relevant knowledge it has acquired, rather than through its surface observations. We realize this abstraction in two steps: constructing a query-specific set of atomic knowledge conditions, and representing each trajectory step by the subset of conditions it has satisfied.

\paragraph{Knowledge conditions.}
For each query $Q$ with reference answer $y^*$, we construct a finite set of atomic knowledge conditions
\begin{equation}
\mathcal{K}(Q) = \{k_1, \dots, k_K\},
\end{equation}
where each $k_j$ is a single-hop factual constraint needed to derive $y^*$. Intuitively, $\mathcal{K}(Q)$ encodes the answer's logical skeleton: a trajectory cannot reliably produce $y^*$ without acquiring every $k_j$. We construct $\mathcal{K}(Q)$ through a two-stage pipeline that first identifies the reasoning skeleton implied by $Q$ (with placeholders for unknown entities) and then grounds each placeholder using $y^*$ and a successful reference trajectory. This decoupling produces a trajectory-agnostic knowledge space and ensures three properties: each condition encodes a single fact (atomicity), the conditions jointly support the full reasoning chain to $y^*$ (completeness), and entity references remain consistent across the query, the answer, and the search trajectories (grounding consistency). Construction details and the prompts used at each stage are deferred to Appendix~\ref{app:knowledge_condition_construction}.

\paragraph{Knowledge state.}
Given $\mathcal{K}(Q)$, we represent the state of a trajectory $\tau_i$ at step $t$ by the subset of conditions it has satisfied so far. At each step, an LLM-based verifier inspects the new observation $o_{i,t}$ and outputs a binary vector $\Delta\mathbf{s}_{i,t} \in \{0,1\}^K$ indicating the set of conditions newly verifiable from $o_{i,t}$:
\begin{equation}
    [\Delta \mathbf{s}_{i,t}]_j = \mathbb{I}\!\left(h_{i,t} \models k_j\right).
\end{equation}
We then monotonically accumulate these increments to obtain the cumulative knowledge state:
\begin{equation}
\mathbf{s}_{i,t} = \mathbf{s}_{i,t-1} \vee \Delta\mathbf{s}_{i,t}, \qquad \mathbf{s}_{i,0} = \mathbf{0}_K,
\end{equation}
where $\vee$ denotes element-wise OR. Monotonic accumulation reflects the fact that acquired knowledge cannot be unacquired, and absorbs transient verifier noise across nearby steps. The knowledge state $\mathbf{s}_{i,t}$ defines a shared semantic space across episodes: two steps with the same $\mathbf{s}$ have, by construction, acquired the same set of task-relevant facts, regardless of which queries were issued, which pages were visited, or how reasoning was phrased in between. This is the property that surface observations fail to provide and that KnowGPO exploits in the next subsection.

\subsection{Knowledge-Grouped Policy Optimization}
\label{sec:knowgpo}

Given the knowledge state representation, Know-GPO assigns credit at two levels. Episode-level advantages align the policy with the final task objective, and step-level advantages, computed by grouping steps under matched knowledge states, attribute credit to individual search decisions.

\paragraph{Episode-level advantage.}
For each query $Q$, we sample $N$ episodes $\{\tau_i\}_{i=1}^N$ from the current policy and compute the group-normalized advantage of episode $i$ as
\begin{equation}
A_i^E = \frac{R(\tau_i) - \mu}{\sigma + \varepsilon},
\end{equation}
where $\mu$ and $\sigma$ are the empirical mean and standard deviation of $\{R(\tau_i)\}_{i=1}^N$, and $\varepsilon$ is a small constant for numerical stability. This is the standard outcome-based group advantage of GRPO.

\paragraph{Step-level advantage via knowledge-grouped comparison.}
To attribute credit at the step-level, we exploit the fact that the knowledge state $\tilde{\mathbf{s}}$ partitions the space of intermediate steps into semantically equivalent classes. Let $\mathcal{U}$ denote the set of distinct knowledge states observed across all $N$ episodes for query $Q$. For each $\tilde{\mathbf{s}} \in \mathcal{U}$, we form a knowledge group
\begin{equation}
\mathcal{G}(\tilde{\mathbf{s}}) = \left\{(i, t) \;\middle|\; \mathbf{s}_{i,t} = \tilde{\mathbf{s}},\; a_{i,t} \neq \texttt{answer}\right\}
\end{equation}
containing every non-terminal step that starts from state $\tilde{\mathbf{s}}$. We then define the step-level reward as the \textit{incremental knowledge gain},
\begin{equation}
r_{i,t} = \|\mathbf{s}_{i,t}\|_1 - \|\mathbf{s}_{i,t-1}\|_1,
\end{equation}
which counts the number of new knowledge conditions satisfied by step $(i, t)$. The step-level advantage is the group-normalized incremental gain within $\mathcal{G}(\tilde{\mathbf{s}})$:
\begin{equation}
A_{i,t}^{S} = \frac{r_{i,t} - \mu_{\tilde{\mathbf{s}}}}{\sigma_{\tilde{\mathbf{s}}} + \varepsilon},
\end{equation}
where $\mu_{\tilde{\mathbf{s}}}$ and $\sigma_{\tilde{\mathbf{s}}}$ are computed over $\mathcal{G}(\tilde{\mathbf{s}})$. Two features of this construction are important. First, all steps compared within a group share the same prior knowledge state, so $A_{i,t}^{S}$ measures how much more (or less) a step contributes to knowledge acquisition than alternative actions taken from the \emph{same} epistemic position. Second, the step-level signal is anchored to the environment through the verifier's checks against $\mathcal{K}(Q)$, rather than to surface observations or the policy's own answer probabilities.

\paragraph{Combined objective.}
For each step $(i, t)$, we combine the two advantages with a single mixing coefficient $\omega \in [0, 1]$:
\begin{equation}
A_{i,t} = (1 - \omega)\, A_i^E + \omega\, A_{i,t}^{S}.
\end{equation}
The episode-level term keeps optimization aligned with the final outcome and prevents the policy from collapsing onto local information-gathering behaviors. The step-level term provides dense, semantically grounded supervision that distinguishes between informative and redundant decisions sharing the same starting knowledge.

We optimize the policy with a PPO-style clipped objective~\citep{schulman2017proximalpolicyoptimizationalgorithms}:
\begin{equation}
\begin{aligned}
\mathcal{L}_{\text{KnowGPO}}(\theta) =\; & \mathbb{E}\!\left[\sum_{i,t} \min\!\Big(\rho_{i,t}\, A_{i,t},\right. \\
& \left. \quad \mathrm{clip}(\rho_{i,t},\, 1-\epsilon,\, 1+\epsilon)\, A_{i,t}\Big)\right] \\
& \;-\; \beta\, D_{\mathrm{KL}}\!\left(\pi_\theta \,\|\, \pi_{\mathrm{ref}}\right),
\end{aligned}
\end{equation}
The full procedure, including knowledge state verification and group construction, is summarized in Algorithm~\ref{alg:knowgpo} (Appendix~\ref{app:algorithm}).

\section{Experiments}

We evaluate KnowGPO on four research questions. First, does knowledge-grouped credit assignment yield consistent gains over outcome-only and process-supervision baselines (Section~\ref{sec:main_results})? Second, are knowledge-state grouping and incremental knowledge gain each individually necessary (Section~\ref{sec:ablation})? Third, does the knowledge state genuinely provide a better grouping anchor than surface observations (Section~\ref{sec:grouping_anchor})? Fourth, how does KnowGPO change search behavior, and at what training cost (Section~\ref{sec:analysis})?

\subsection{Experimental Setup}
\paragraph{Benchmarks.}
We evaluate KnowGPO on three challenging deep search benchmarks that cover diverse question distributions and difficulty levels: Xbench~\citep{xbench2025deepsearch} (covering the 2505 and 2510 splits), BrowseComp-ZH~\citep{zhou2025browsecompzh}, and the text-only subset of GAIA~\citep{mialon2023gaia}. Following standard LLM-as-judge protocols, GPT-4o~\citep{openai2024gpt4ocard} judges whether the agent's final response matches the ground truth. All evaluations follow each benchmark's official protocol; we report the mean accuracy over 3 independent runs.

\paragraph{Baselines.}
We compare KnowGPO against three representative reinforcement learning approaches, all trained from the same SFT checkpoint with identical data and tool environments. \textbf{GRPO}~\citep{shao2024deepseekmathpushinglimitsmathematical} uses only the binary outcome reward and is the dominant baseline in current open-source systems. \textbf{GiGPO}~\citep{feng2025groupingrouppolicyoptimizationllm} introduces step-level credit assignment by grouping steps that revisit the same intermediate state and computing within-group advantages from discounted future returns. \textbf{IGPO}~\citep{wang2026informationgainbasedpolicyoptimization} assigns each step a dense reward equal to the increase in the model's predicted probability of the correct answer. In addition, we report numbers from several state-of-the-art open-source and proprietary deep search agents for context; differences in training data, context length, and tool stacks limit direct comparison, and we mark such systems separately.

\paragraph{Implementation details.}
All trained models use Qwen3-8B~\citep{yang2025qwen3technicalreport} as the base model, with context extended to 128K tokens via YaRN~\citep{peng2026yarnefficientcontextwindow}. SFT is conducted with LlamaFactory~\citep{zheng2024llamafactoryunifiedefficientfinetuning}; RL training uses verl~\citep{Sheng_2025}. All baselines share the same SFT initialization, training data, learning rate, batch size, and rollout configuration as KnowGPO; only the advantage estimator and reward design differ. Knowledge state verification uses DeepSeek-v3.2 as the LLM verifier; the choice and its quality analysis are reported in Appendix~\ref{app:verifier-selection}. Full hyperparameters are in Appendix~\ref{app:implementation}.

\begin{table}[t]
  \centering
  \small
  
  \begin{tabular}{l c c}
  \toprule
  \textbf{Model} & \textbf{BrowseComp-ZH} & \textbf{GAIA} \\
  \midrule
  
  \rowcolor[HTML]{F2F2F2}
  \multicolumn{3}{l}{\textit{Advanced Agents with Proprietary Data}} \\
  MiniMax-M2.1        & 47.8 & 64.3 \\
  DeepSeek-V3.1       & 49.2 & 63.1 \\
  DeepSeek-R1         & 26.3 & 16.5 \\
  GLM-4.5             & 37.5 & 66.0 \\
  Kimi-K2.5           & 62.3 & --   \\
  Claude-4-Sonnet     & 29.1 & 68.3 \\
  \midrule
  
  \rowcolor[HTML]{F2F2F2}
  \multicolumn{3}{l}{\textit{Agents with Open-source Data}} \\
  WebExplorer-8B      & 32.0 & 50.0 \\
  WebSailor-7B        & 14.2 & 37.9 \\
  WebSailor-32B       & 25.5 & 51.5 \\
  ASearcher-Web-32B   & 15.6 & 52.8 \\
  DeepMiner-32B-RL    & 40.1 & 58.7 \\
  DeepDive-9B         & 15.1 & --   \\
  DeepDive-32B        & 29.7 & --   \\
  \midrule
  
  \rowcolor[HTML]{F2F2F2}
  \multicolumn{3}{l}{\textit{Our Agents}} \\
  Qwen3-8B-SFT        & 30.1 & 31.1 \\
  \quad + GRPO        & 32.9 & 35.0 \\
  \quad + GiGPO       & \underline{37.3} & 35.0 \\
  \quad + IGPO       & 32.2 & \underline{37.9} \\
  \quad + \textbf{KnowGPO}
                      & \textbf{38.8} & \textbf{40.2} \\
  \bottomrule
  \end{tabular}
  \caption{\textbf{Performance comparison on BrowseComp-ZH and GAIA.} Among RL-based methods, best results are in \textbf{bold}, and second-best are \underline{underlined}.}
  \label{tab:results_part1}
\end{table}

\begin{table}[t]
  \label{tab:results_part2}
  \centering
  \small
  
  \begin{tabular}{l c c}
  \toprule
  \textbf{Model} & \textbf{Xbench-2505} & \textbf{Xbench-2510} \\
  \midrule
  
  \rowcolor[HTML]{F2F2F2}
  \multicolumn{3}{l}{\textit{Advanced Agents with Proprietary Data}} \\
\  MiniMax-M2.1        & 68.7 & 43.0 \\
  DeepSeek-V3.2       & 78.0 & 55.7 \\
  GLM-4.7             & 72.0 & 52.3 \\
  Kimi-K2.5           & --   & 46.0 \\
  Claude-4-Sonnet     & 64.6 & --   \\
  OpenAI-o3           & 67.0 & -- \\
  OpenAI GPT-5-High   & 77.8 & 75.0 \\
  Gemini-3.0-Pro      & --   & 53.0 \\
  \midrule
  
  \rowcolor[HTML]{F2F2F2}
  \multicolumn{3}{l}{\textit{Agents with Open-source Data}} \\
  WebExplorer-8B    & 53.7 & 23.0 \\
  MiroThinker-v1.0-8b  & 60.6 & 34.0 \\
  MiroThinker-v1.0-30B & 70.6 & --   \\
  MiroThinker-v1.5-30B & 73.1 & --   \\
  MiroThinker-v1.7-mini  & --   & 57.2 \\
  \midrule
  
  \rowcolor[HTML]{F2F2F2}
  \multicolumn{3}{l}{\textit{Our Agents}} \\
  Qwen3-8B-SFT        & 39.0 & 29.0 \\
  \quad + GRPO        & \underline{46.0} & 34.0 \\
  \quad + GiGPO       & 44.0 & \underline{35.0} \\
  \quad + IGPO       & \underline{46.0} & 34.0 \\
  \quad + \textbf{KnowGPO}
                      & \textbf{48.0} & \textbf{36.0} \\
  \bottomrule
  \end{tabular}
    \caption{\textbf{Performance comparison on Xbench-2505 and Xbench-2510.} Among RL-based methods, best results are in \textbf{bold}, and second-best are \underline{underlined}.}
    \label{tab:results_part2}
\end{table}

\subsection{Main Results}
\label{sec:main_results}

Tables~\ref{tab:results_part1} and~\ref{tab:results_part2} report performance on all three benchmarks. We make two observations.

\paragraph{KnowGPO consistently outperforms both outcome-only and step-level baselines.}
All RL approaches improve over SFT, confirming that explicit policy optimization is necessary for deep search agents. However, existing step-level methods show considerable cross-dataset variability: GiGPO degrades by 2.0\% on Xbench-2505 relative to GRPO despite winning on BrowseComp-ZH, and IGPO matches GRPO on Xbench-2505 but barely improves on BrowseComp-ZH. KnowGPO is the only method that improves on every split, gaining 5.9\% on BrowseComp-ZH and 5.2\% on GAIA over GRPO, and exceeding the strongest step-level baseline by 1.5\% and 2.3\% on these two benchmarks. The averaged gain over the SFT initialization is 8.5\%. We attribute this consistency to the fact that knowledge-grouped credit signals do not depend on either surface state recurrence or stable model confidence, both of which are particularly fragile on noisier benchmarks.

\paragraph{KnowGPO closes the gap to 32B open-sour-ce deep search agents at 8B scale.}
On BrowseComp-ZH, KnowGPO reaches 38.8\%, surpassing the same-scale WebExplorer-8B by 6.8\% and the 32B WebSailor-32B and DeepDive-32B by 13.3\% and 9.1\% respectively, and trailing the strongest open-source system DeepMiner-32B-RL by only 1.3\%. On Xbench-2505 it reaches 48.0\%, outperforming the 8B MiroThinker-v1.0 by 12.6\%. These results show that knowledge-grouped credit assignment narrows the gap to substantially larger open-source systems without proprietary data or larger backbones. A gap to proprietary systems remains, for example 23.5\% to Kimi-K2.5 on BrowseComp-ZH and 28.1\% to Claude-4-Sonnet on GAIA, which we attribute to training data scale, model capacity, and inference-time optimization beyond our scope.

\begin{figure}[] 
  \centering
  \includegraphics[width=1.0\linewidth]{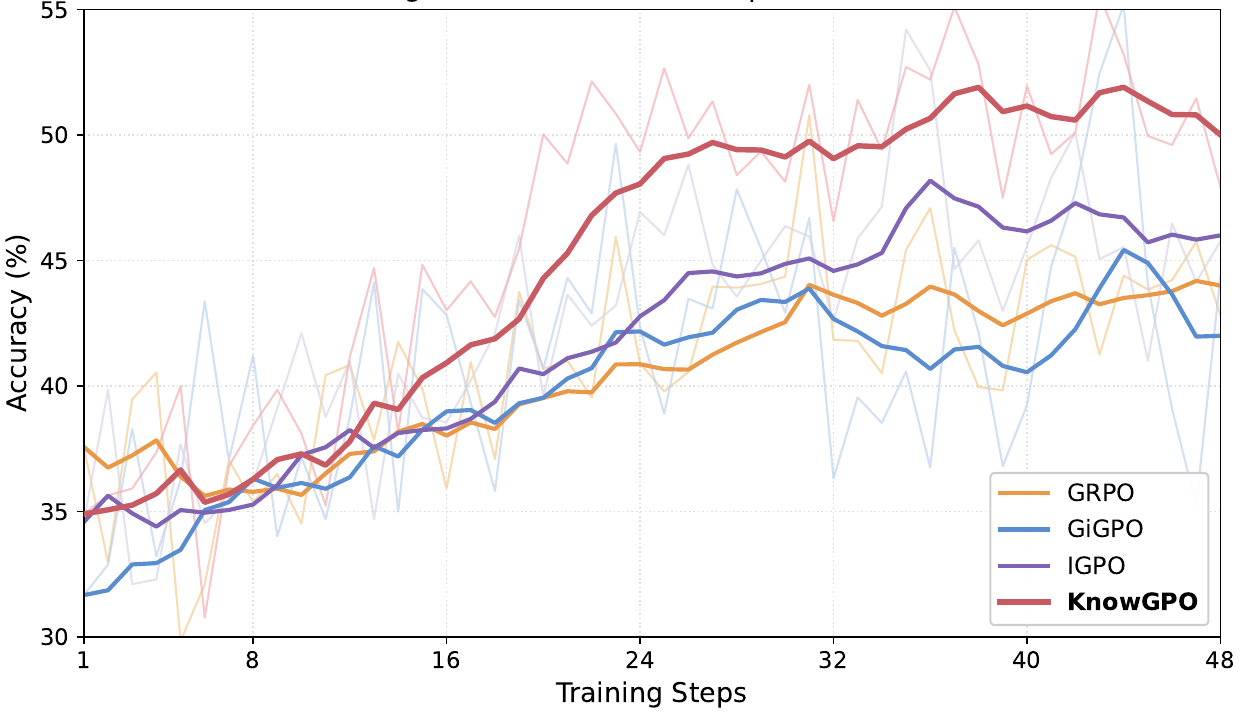} 
  \caption{Training curves comparing GRPO, GiGPO, IGPO, and KnowGPO on our Qwen3-8B-SFT.}
  \label{fig:training_curves}
\end{figure}

\paragraph{Training dynamics across baselines.}
Figure~\ref{fig:training_curves} tracks validation accuracy on Xbench-2505 in RL training. All four methods start from the same SFT checkpoint and follow comparable trajectories during the first 16 steps, after which they diverge. 
KnowGPO opens a clear gap by around step 20 and plateaus near step 32 at roughly 50\%, while GRPO, GiGPO, and IGPO settle below 46\%. 
The gap is consistent across the second half of training, suggesting that knowledge-grouped credit assignment provides a denser signal throughout rather than a small final-stage advantage.

\subsection{Ablation Study}
\label{sec:ablation}

We next ablate the components of KnowGPO. The first ablation studies the mixing coefficient $\omega$; the second isolates the contributions of incremental knowledge gain and knowledge-state grouping; the third compares incremental gain against discounted future returns as a process reward. All ablations use the same SFT initialization and training setup as the main experiments.

\begin{table}[t]
  \centering
  \small
  \setlength{\tabcolsep}{3pt}  
  
  \begin{tabular}{l|c c c c}  
  \toprule
  \textbf{Model} & \textbf{Xben05} & \textbf{Xben10} & \textbf{BC-ZH} & \textbf{GAIA} \\
  \midrule
  Qwen3-8b-SFT              & 39.0 & 29.0 & 30.1 & 31.1 \\
  \;+KnowGPO ($\omega{=}0$)     & 45.0 & 28.0 & 34.0 & 34.0 \\
  \rowcolor[HTML]{F2F2F2} \;+KnowGPO ($\omega{=}0.3$)   & \textbf{48.0} & \textbf{36.0} & \textbf{38.8} & \textbf{40.2} \\
  \;+KnowGPO ($\omega{=}0.5$)   & 47.0 & 35.0 & 36.3 & 37.9 \\
  \bottomrule
  \end{tabular}
  \caption{Performance of KnowGPO with different $\omega$.}
  \label{tab:omega_ablation}
\end{table}
\begin{table}[t] 
  \centering
  \footnotesize 
  \setlength{\tabcolsep}{3.5pt} 

  \begin{tabular}{l|cccc}
  \toprule
  \textbf{Method} & \textbf{Xben05} & \textbf{Xben10} & \textbf{BC-ZH} & \textbf{GAIA} \\
  \midrule
  GRPO (Base)              & 46.0 & 34.0 & 32.9 & 35.0 \\
  \quad + $\Delta$ KG  & 47.0 & 34.0 & 34.6 & 37.8 \\
  KnowGPO (w/ $G_t$)       & 45.0 & 35.0 & 32.8 & 36.8 \\
  \rowcolor[HTML]{F2F2F2} \textbf{KnowGPO}  & \textbf{48.0} & \textbf{36.0} & \textbf{38.8} & \textbf{40.2} \\
  \bottomrule
  \end{tabular}
  \caption{\textbf{Ablation study on reward design and grouping mechanisms.} (1) GRPO with outcome rewards, (2) GRPO with incremental knowledge gain, (3) KnowGPO with discounted future returns, and (4) KnowGPO.}
  \label{tab:ablation_study}
\end{table}

\paragraph{The mixing coefficient $\omega$ balances outcome and process supervision.}
Table~\ref{tab:omega_ablation} sweeps $\omega \in \{0, 0.3, 0.5\}$. Performance peaks at $\omega = 0.3$ across all four splits. Setting $\omega = 0$ disables step-level supervision yet still outperforms GRPO on BrowseComp-ZH and GAIA, suggesting that grouping by knowledge state implicitly regularizes learning even before any step reward is applied. Setting $\omega = 0.5$ overweights process supervision and degrades on three of four splits, reflecting that excessive local optimization can pull the policy away from the terminal objective. We use $\omega = 0.3$ throughout the rest of the paper.

\paragraph{Both incremental gain and knowledge-state grouping are individually necessary.}
Table~\ref{tab:ablation_study} decomposes KnowGPO against GRPO. Adding incremental knowledge gain to GRPO (\textit{+$\Delta$KG}, applied without knowledge-state grouping) improves over GRPO by 1.0\% to 2.8\% across benchmarks, confirming that knowledge-grounded step rewards are useful even without proper grouping. Further introducing knowledge-state grouping (KnowGPO, full) yields an additional 1.0\% to 4.2\%, with the largest gain (+4.2\%) on BrowseComp-ZH. The role of grouping is to restrict normalization to steps that share the same prior knowledge state, so that within-group comparison contrasts decisions made from the same epistemic position rather than across mixed positions; the ablation isolates this effect from the choice of process reward.

\paragraph{Incremental knowledge gain outperforms discounted future returns as a process reward.}
Within the same knowledge-grouped framework, replacing incremental gain with the discounted future return $G_t$ used by GiGPO drops accuracy by 3.0\%, 6.0\%, and 3.4\% on Xbench-2505, BrowseComp-ZH, and GAIA respectively (Table~\ref{tab:ablation_study}, row "KnowGPO w/ $G_t$"). The two rewards differ in how much downstream stochasticity they absorb: $G_t$ couples a step's credit to all subsequent decisions and is sensitive to long-horizon noise, while incremental knowledge gain is a local, immediate signal that does not require waiting for the trajectory to end. The gap shows that, even when grouping is fixed, the choice of step reward materially matters for long-horizon credit assignment.

\subsection{Why Knowledge-Based Grouping Works}
\label{sec:grouping_anchor}

The ablation shows that knowledge-state grouping matters in aggregate. We now examine \emph{why} it works by directly measuring the grouping behavior of three candidate anchors on real rollouts.

\paragraph{Setup.}
We sample 500 queries from Xbench and BrowseComp~\citep{wei2025browsecomp} and rollout the SFT checkpoint up to 16 times per query, producing 5,922 valid trajectories with 55,571 steps after filtering. We compare three grouping anchors: \textit{observation-anchored}, matching steps by BGE-large cosine similarity over retrieved contents at threshold 0.9 (the protocol used by GiGPO); \textit{action-anchored}, matching steps by similarity of generated queries; and \textit{knowledge-anchored} (ours), matching steps by exact equality of accumulated knowledge state. Grouping quality is measured by the effective step ratio (ESR, fraction of steps in non-singleton groups), the singleton ratio, the mean non-singleton group size, and the cross-rollout coverage (average number of distinct rollouts per non-singleton group). Formal definitions are in Appendix~\ref{app:gigpo-limitations}.

\paragraph{Knowledge state aligns far more steps than observation or action.}

\begin{table}[t]
\centering
\small 
\begin{tabular*}{\columnwidth}{l@{\extracolsep{\fill}}ccc}
\toprule
Metric & Obs & Act & Know \\
\midrule
ESR $\uparrow$ (\%) & 51.8 & 79.4 & \textbf{97.8} \\
SR $\downarrow$ (\%) & 79.4 & 72.0 & \textbf{19.0} \\
MeanNG  & 4.85 & 24.62 & 21.28 \\
Cross $\uparrow$ & 3.83 & \textbf{4.85} & 4.66 \\
$n_{\mathrm{groups}}$ & 32,267 & 14,902 & \textbf{4,872} \\
\bottomrule
\end{tabular*}
\caption{Comparison of grouping quality across three state anchors.}
\label{tab:diag-main}
\end{table}


Table~\ref{tab:diag-main} reports the five metrics. Observation-based grouping, the anchor used by GiGPO, leaves only 51.8\% of steps inside non-singleton groups and produces 79.4\% singletons, which means most intended step-level comparisons cannot be executed: the within-group advantage degenerates to comparing a step with itself, providing no learning signal. Action-based grouping improves ESR to 79.4\% but its singleton ratio is still 72.0\%, indicating that even when two trajectories issue similar queries, the evidence they end up acquiring is rarely comparable. Knowledge-based grouping raises ESR to 97.8\% and reduces singletons to 19.0\%, while keeping cross-rollout coverage (4.66) close to the highest value across anchors. This last metric is important: it confirms that the gain is not an artifact of merging adjacent steps within the same rollout, but reflects alignment of semantically equivalent steps across independent trajectories. A qualitative case in Appendix~\ref{app:gigpo-limitations} shows two steps with 0.987 retrieval similarity that are correctly separated by the knowledge anchor (one has acquired 0 of 12 conditions, the other 8 of 12), illustrating that surface similarity can hide drastic differences in actual search progress.

\subsection{Search Behavior and Training Efficiency}
\label{sec:analysis}


\begin{table}[t]
  \centering
  \scriptsize
  \setlength{\tabcolsep}{3pt}
  \renewcommand{\arraystretch}{1.0}
  \resizebox{\linewidth}{!}{
  \begin{tabular}{l|cccc}
  \toprule
  \textbf{Method} & \textbf{Both Fail} & \textbf{Both Pass} & \textbf{GRPO-only} & \textbf{Know-only} \\
  \midrule
  GRPO    & 11.39 & 7.03 & 9.51  & 11.55 \\
  KnowGPO & 10.85 & 6.33 & 10.36 & \textbf{9.83} \\
  \bottomrule
  \end{tabular}
  }
  \caption{\textbf{Average search turns across outcome subsets.} Both Fail: queries where both methods fail; Both Pass: both succeed; GRPO-only: only GRPO succeeds; KnowGPO-only: only KnowGPO succeeds.}
  \label{tab:search_efficiency}
\end{table}
\begin{figure}[t] 
  \centering
  \includegraphics[width=1.0\linewidth]{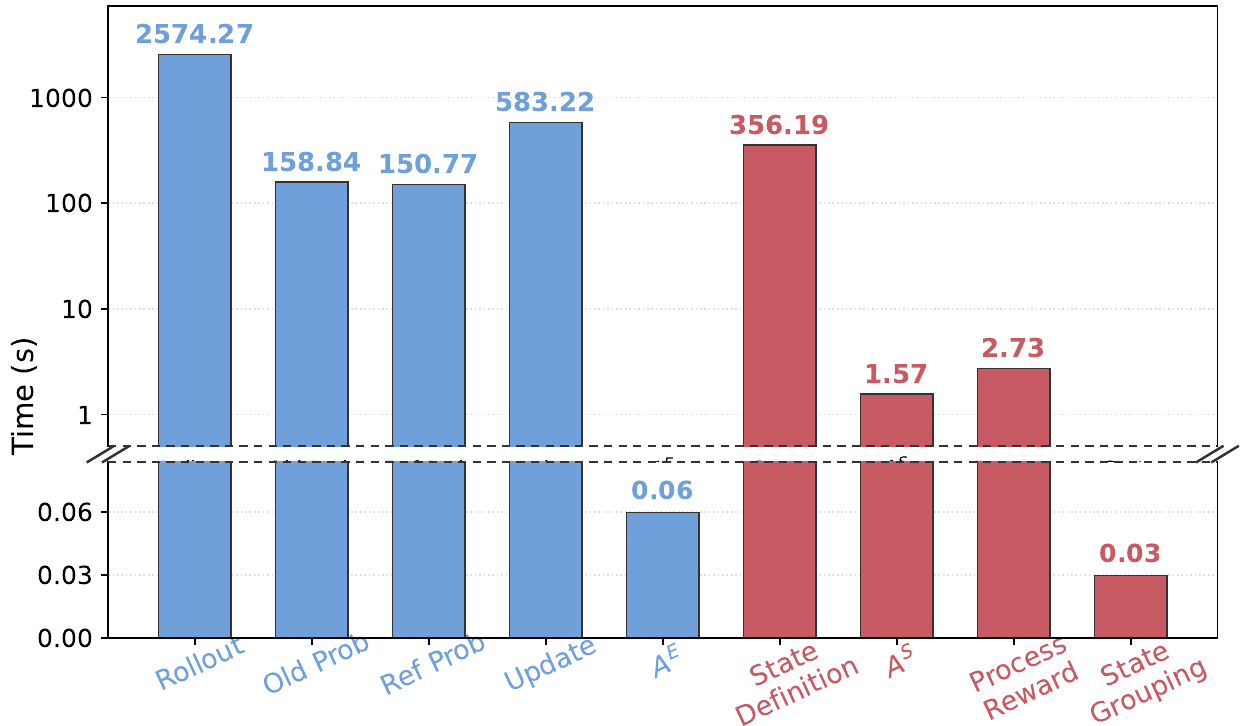} 
  \caption{\textbf{Per-iteration training time breakdown.} Blue bars represent shared RL operations common to both GRPO and KnowGPO; red bars represent KnowGPO-specific components.}
  \label{fig:training_effiency}
\end{figure}

\paragraph{KnowGPO learns to search adaptively, not just less.}
Reducing redundant search is easy if the policy simply stops earlier on every query, but this would harm hard cases. To test for this failure mode, we partition 500 test queries by joint solving outcome and compare average search turns (Table~\ref{tab:search_efficiency}). On queries solved exclusively by KnowGPO, KnowGPO uses 9.83 turns versus 11.55 for GRPO, a clear reduction in redundant exploration. On queries solved only by GRPO, KnowGPO still uses 10.36 turns, and on queries where both methods fail, it uses 10.85, the longest of all subsets. The agent does not uniformly shorten search; it shortens search when evidence becomes sufficient and continues searching when evidence remains scarce. A representative trajectory comparison in Appendix~\ref{app:search} shows GRPO overshooting a correct intermediate answer by four additional turns and overwriting it with a wrong one, a failure mode KnowGPO avoids.

\paragraph{Knowledge-grouped credit adds negligible training cost.}
A natural concern with structured process supervision is overhead. We profile per-iteration time over three training epochs (Figure~\ref{fig:training_effiency}). The training budget is dominated by operations shared with GRPO: trajectory generation (2,574s), policy update (583s), and probability computation (310s). KnowGPO-specific operations are knowledge verification (356s), incremental gain computation (2.73s), and group construction (0.03s). Verification runs asynchronously and overlaps with the next rollout, so its wall-clock cost is hidden behind generation; the two pure-bookkeeping operations together account for less than 0.08\% of the budget. Knowledge-grouped credit assignment is therefore essentially free compared to GRPO.

\section{Conclusion}

We presented KnowGPO, a reinforcement learning framework that assigns credit by aligning deep search trajectories through the task-relevant knowledge they have acquired, rather than through their surface observations. KnowGPO groups steps under matched knowledge states and uses incremental knowledge gain as a process reward grounded in the environment. Across Xbench, BrowseComp-ZH, and GAIA, KnowGPO improves over outcome-only and process-supervision baselines by 3.8\% to 4.2\%, with an 8B agent matching 32B open-source systems at training cost comparable to GRPO.

\clearpage

\section*{Limitations}

KnowGPO assumes that progress in the target task can be characterized by a finite set of verifiable knowledge conditions, which fits knowledge-intensive search but not more open-ended agentic tasks whose progress is harder to discretize. The fidelity of the knowledge state also depends on an LLM verifier, and although we use a strong external model with monotonic accumulation to suppress transient noise, residual errors can still propagate into step-level advantages. Our construction pipeline additionally requires at least one successful reference trajectory per query to ground entity placeholders, which limits applicability to settings where successful rollouts are reachable at reasonable cost. Finally, all experiments use Qwen3-8B; whether the effects we observe transfer to substantially smaller or larger backbones, or to reasoning-trained initializations, remains to be verified.



\section*{Acknowledgment}

This work was supported by the Fundamental and Interdisciplinary Disciplines Breakthrough Plan of the Ministry of Education of China (JYB2025XDXM103), National Natural Science Foundation of China (No. 62506332) , the Key Research and Development Program of Zhejiang Province, China (No. 2026C02A1240) and CCF-Tencent Rhino-Bird Open Research Fund.

\bibliography{custom}

\clearpage
\appendix

\section{Algorithm Details}
\label{app:algorithm}

This appendix provides the full pseudocode of KnowGPO and the explicit training objective used for parameter updates. The procedure begins by sampling $N$ trajectories for each query and running the LLM verifier at every step to construct the knowledge state $\mathbf{s}_{i,t}$. Trajectories are then grouped by their preceding knowledge state, and step-level advantages are computed by within-group normalization of incremental knowledge gain. The episode-level outcome advantage and the step-level knowledge-grouped advantage are combined with a single mixing coefficient $\omega$ and used as the per-token advantage in a PPO-style clipped objective.

\definecolor{algcomment}{RGB}{34, 139, 34}
\newcommand{\algcomment}[1]{\textcolor{algcomment}{\(\triangleright\) \textit{#1}}}

\begin{algorithm}[]
\caption{Knowledge-Grouped Policy Optimization (KnowGPO)}
\label{alg:knowgpo}
\begin{algorithmic}[1]
\Require Query $Q$, reference $y^*$, conditions $\mathcal{K}(Q)$,policy $\pi_\theta$, group size $N$, mixing $\omega$
\Statex \algcomment{Roll out and accumulate knowledge state}
\State Sample $N$ trajectories $\{\tau_{i}\}_{i=1}^N \sim \pi_\theta(\cdot \mid Q)$
\For{each trajectory $\tau_{i}$ and step $t$}
    \State $\Delta \mathbf{s}_{i,t} \gets \text{Verifier}(o_{i,t}), \mathcal{K}(Q))$
    \State $\mathbf{s}_{i,t} = \mathbf{s}_{i,t-1} \vee \Delta\mathbf{s}_{i,t}$
\EndFor
\Statex \algcomment{Episode-level advantage from outcome}
\State $A_i^{\mathrm{E}} \gets \text{Normalize}\bigl(R(\tau_{i})\bigr)$
    \Statex \hspace{1.4em} over $\{R(\tau_{j})\}_{j=1}^N$
\Statex \algcomment{Group steps by matched knowledge state}
\For{each distinct $\tilde{\mathbf{s}}$ observed}
    \State $\mathcal{G}(\tilde{\mathbf{s}}) \gets \{(i,t) : \mathbf{s}_{i,t-1} = \tilde{\mathbf{s}},$
        \Statex \hspace{2.6em} $a_{i,t} \neq \texttt{answer}\}$
    \Statex \hspace{1.4em} \algcomment{Incremental gain as step reward}
    \State $r_{i,t} \gets \|\mathbf{s}_{i,t}\|_1 - \|\mathbf{s}_{i,t-1}\|_1$
    \State $A_{i,t}^{\mathrm{S}} \gets \text{Normalize}(r_{i,t})$ over $\mathcal{G}(\tilde{\mathbf{s}})$
\EndFor
\Statex \algcomment{Combine advantages and update policy}
\State $A_{i,t} \gets (1-\omega) A_i^{\mathrm{E}} + \omega A_{i,t}^{\mathrm{S}}$
\State Update $\theta$ by maximizing $\mathcal{L}_{\text{KnowGPO}}(\theta)$
\end{algorithmic}
\end{algorithm}

\paragraph{Token-level advantage broadcast.}
Within each step $(i,t)$, the advantage $A_{i,t}$ is broadcast uniformly across all tokens that compose the action $a_{i,t}$ and its preceding reasoning trace, following standard practice in GRPO-style training. The clipping ratio $\epsilon$, KL coefficient $\beta$, and reference policy $\pi_{\mathrm{ref}}$ are inherited from the SFT checkpoint and are shared across all baselines we compare against.

\paragraph{Asynchronous verifier execution.}
The verifier call on line 4 of Algorithm~\ref{alg:knowgpo} is the heaviest operation introduced by KnowGPO. We run it asynchronously, in parallel with the next rollout batch, so that its wall-clock cost is hidden behind trajectory generation. Empirical timing in Section~\ref{sec:analysis} confirms this overlap makes verification effectively free relative to GRPO.

\section{Knowledge Condition Construction}
\label{app:knowledge_condition_construction}

This appendix details the two-stage pipeline introduced in Section~\ref{sec:knowledge_state} for constructing the knowledge condition set $\mathcal{K}(Q)$.

\paragraph{Three design principles.}
We require $\mathcal{K}(Q)$ to satisfy three properties.
(i) \textbf{Atomicity}: each $k_j$ encodes a single-hop factual constraint, avoiding compound multi-hop statements inside a single condition.
(ii) \textbf{Completeness}: the set jointly supports the full reasoning chain from $Q$ to $y^*$, so that omitting any $k_j$ renders $y^*$ non-derivable.
(iii) \textbf{Grounding consistency}: entities in each condition use consistent referential forms across $Q$, $y^*$, and search trajectories, so that hit determination is not destabilized by entity ambiguity.

\paragraph{Bipartite graph view.}
We represent $\mathcal{K}(Q)$ as a constraint-entity bipartite graph
\begin{equation}
\begin{aligned}
G &= (V_C \cup V_E, E),\\
V_C &= \{C_1, \dots, C_m\},\\
V_E &= \{\langle E_0 \rangle, \langle E_1 \rangle, \dots\},
\end{aligned}
\end{equation}
where $\langle E_0 \rangle$ is the target answer entity and an edge connects $C_i$ to $\langle E_j \rangle$ whenever $C_i$ explicitly contains $\langle E_j \rangle$. We require $G$ to be connected under BFS from $\langle E_0 \rangle$, which ensures every entity in $\mathcal{K}(Q)$ traces back to the answer through factual constraints and rules out isolated knowledge.

\paragraph{Two-stage construction.}
\emph{Stage 1 (Query $\to$ Placeholder Knowledge).} The first stage takes only $Q$ as input and outputs placeholder conditions that capture the reasoning skeleton implicit in the query. Entities are denoted by placeholders $\langle E_0\rangle, \langle E_1\rangle, \dots$, where $\langle E_0\rangle$ is fixed as the target answer entity. This stage is trajectory-agnostic and does not require access to $y^*$.

\emph{Stage 2 (Placeholder Knowledge $\to$ Grounded Knowledge).} The second stage takes $Q$, $y^*$, the Stage 1 placeholders, and a successful reference trajectory obtained via rejection sampling. The model grounds each placeholder $\langle E_j \rangle$ to its real-world entity, producing the fully grounded $\mathcal{K}(Q)$ used at training time.

Decoupling structural decomposition (Stage 1) from entity grounding (Stage 2) yields a trajectory-agnostic knowledge space whose structure does not depend on which particular rollout happens to succeed. The prompts used at each stage are shown below.

\begin{examplebox}{Prompt Template for Stage 1}
You are an expert in logical reasoning and knowledge decomposition. 
You will receive a complex **Deep Search question**. Let <E0> be the 
final **answer entity** to the question.\par

Your task is to break down the question into a list of **knowledge 
conditions** that <E0> should satisfy.\par

\#\# Requirements for the Knowledge Conditions\par
1. Each knowledge condition must be a **single-hop factual statement**, 
   where the intermediate entities should be denoted as <E1>, <E2>, 
   <E3>, and so on.\par
2. Each knowledge condition must contain **at least one entity 
   placeholder**.\par
3. Each statement must be **clear, coherent, and grammatically correct**.\par
4. Do **NOT** attempt to infer or guess the actual identities of any 
   entities.\par
5. **<E0> MUST appear** in at least one knowledge condition.\par
6. The entity graph formed by the knowledge conditions must be 
   **connected**: starting from <E0>, you must be able to reach every 
   other entity through the knowledge relationships (BFS/traversal 
   from <E0> can visit all entity nodes).\par
7. Use the **same language** as the question.\par

\#\# Output Format\par
Your output must follow this format exactly:\par
```\par
[Begin of Knowledge Conditions]\par
\{\{knowledge condition 1\}\}\par
\{\{knowledge condition 2\}\}\par
\{\{knowledge condition 3\}\}\par
...\par
[End of Knowledge Conditions]\par
```\par

\#\#\# Example\par
**Question:**\par
Identify the populated locale defined by the simultaneous 
satisfaction of these constraints: It exists within a jurisdiction 
whose seat of governance is an urban entity named for a major 
settlement in the northern third of a certain European island nation. 
This containing jurisdiction is itself entirely within a larger 
contiguous socio-economic zone. The name of that zone when all such 
zones in its overarching political division are ordered by the number 
of human inhabitants occupies the fifth position. The locale's common 
designation is the answer.\par

**Knowledge Conditions:**\par
```\par
[Begin of Knowledge Conditions]\par
<E0> is a populated locale.\par
<E0> exists within a jurisdiction <E1>.\par
<E1> has a seat of governance that is an urban entity <E2>.\par
<E2> is named for a major settlement in the northern third of a 
certain European island nation <E3>.\par
<E1> is entirely within a larger contiguous socio-economic zone <E4>.\par
<E4> is one of multiple such zones in an overarching political 
division <E5>.\par
When all such zones in <E5> are ordered by the number of human 
inhabitants, <E4> occupies the fifth position.\par
[End of Knowledge Conditions]\par
```\par

\#\# Now, list the knowledge conditions for the following question:\par

[Begin of Question]\par
\{question\}\par
[End of Question]\par

\end{examplebox}
\begin{examplebox}{Prompt Template for Stage 2}
You are an expert entity extractor. You will receive:\par
1. A complex multi-hop **question**\par
2. The correct **answer**\par
3. A set of **knowledge conditions** (with entity placeholders 
   like <E0>, <E1>, <E2>, ...)\par
4. **Thinking** traces and **retrieved documents** from a search 
   agent's trajectory\par

Your task is to identify the real-world entities corresponding to 
each placeholder in the knowledge conditions.\par

\#\# Instructions\par
1. <E0> is always the **final answer entity**.\par
2. For each other placeholder (<E1>, <E2>, ...), determine the 
   actual entity based on the question text, the answer, the 
   thinking traces, and retrieved documents.\par
3. If an entity cannot be confidently determined, use "UNKNOWN" 
   as its value.\par
4. Return ALL entities mentioned in the knowledge conditions, not 
   just the ones you are confident about.\par

\#\# Output Format\par
Return a JSON object exactly like this:\par
```json\par
\{\par
  "entities": \{\par
    "<E0>": "the final answer",\par
    "<E1>": "entity 1 name",\par
    "<E2>": "entity 2 name",\par
    ...\par
  \}\par
\}\par
\#\#\# Example\par
**Question:** Start with a rural settlement in an Asian province, 
birthplace of a notable female activist. An educational facility 
in the same administrative division, which transitioned to a 
secondary school level during a period of widespread international 
conflict, educated a male activist. This male activist, an early 
prominent member of a political movement, later entered into 
matrimony with the aforementioned female activist while both were 
in a European nation. The female activist, herself an early adherent 
to this movement and a leader in its female-focused section, 
eventually engaged in information dissemination work for associated 
labor bodies in a significant urban center. This political movement's 
central institution for ideological guidance and information control, 
founded in the fifth month of a year in the early 1920s, initially 
included five specific operational units. What was the name of the 
unit dedicated to managing current affairs and media reports?\par

**Answer:** News and Public Opinion Unit\par

**Knowledge Conditions:**\par
- <E1> is a rural settlement in an Asian province, <E2>.\par
- <E1> is the birthplace of a notable female activist <E3>.\par
- <E6> married <E3> while both were in a European nation <E8>.\par
- <E7> had a central institution <E12> for ideological guidance 
  and information control.\par
- <E12> initially included five specific operational units, 
  including <E0>.\par
- <E0> was a unit dedicated to managing current affairs and 
  media reports.\par

**Thinking trace:**<Think>xxxxx</Think>

**Retrieved Documents:**<Observation>xxxxx</Observation>

**Output:**\par
```json\par
\{\par
  "entities": \{\par
    "<E0>": "News and Public Opinion Unit",\par
    "<E1>": "Yiwu",\par
    "<E2>": "Zhejiang Province",\par
    "<E3>": "Xiang Jingyu",\par
    "<E6>": "Cai Hesen",\par
    "<E7>": "Communist Party of China",\par
    "<E8>": "France",\par
    "<E12>": "Publicity Department of the CPC Central Committee"\par
  \}\par
\}\par
```\par
\#\# Now extract entities for the following:\par

[Begin of Question]\par
\{question\}\par
[End of Question]\par

[Begin of Answer]\par
\{answer\}\par
[End of Answer]\par

[Begin of Knowledge Conditions]\par
\{knowledges\}\par
[End of Knowledge Conditions]\par

[Begin of Thinking Traces]\par
\{thinking\}\par
[End of Thinking Traces]\par

[Begin of Retrieved Documents]\par
\{docs\}\par
[End of Retrieved Documents]\par
\end{examplebox}

\section{Knowledge State Verifier}
\label{app:verifier-selection}

This appendix justifies our choice of LLM verifier and shows representative behavior. We compared five candidate verifiers on 200 held-out query-trajectory pairs with manually annotated ground truth: DeepSeek-v3.2~\citep{deepseekai2025deepseekv32pushingfrontieropen}, DeepSeek-v3.1~\citep{deepseek2025v31}, DeepSeek-v3.1-Terminus~\citep{deepseek2025v31terminus}, Gemini-2.5-Flash-Lite~\citep{comanici2025gemini25pushingfrontier}, and GPT-4o-mini. Each verifier inspects the cumulative interaction history and outputs, for every condition $k_j \in \mathcal{K}(Q)$, whether it is supported by the trajectory so far.

\paragraph{Selection metrics.}
We report three metrics: \emph{Regression rate}, the fraction of state updates that violate monotonicity (a knowledge condition flips from satisfied back to unsatisfied at a later step); \emph{Parse rate}, the fraction of verifier outputs that successfully parse as valid JSON; and \emph{Agreement}, the fraction of per-condition judgments that match human annotation. Table~\ref{tab:verifier_comparison} shows DeepSeek-v3.2 leading on all three metrics, with 94.7\% agreement and a 5.2\% regression rate. We adopt it as the verifier throughout the paper.

\begin{table}[t]
\centering
\small
\begin{tabular}{lccc}
\toprule
\textbf{Model} & \textbf{Regr.$\downarrow$} & \textbf{Parse} & \textbf{Agree.} \\
\midrule
DeepSeek-v3.2          & \textbf{0.052} & \textbf{0.987}          & \textbf{0.947} \\
Gemini-2.5-Flash-Lite  & 0.057          & 0.971 & 0.746 \\
DeepSeek-v3.1-Terminus & 0.057          & 0.924 & 0.835 \\
DeepSeek-v3.1          & \textbf{0.052} & 0.921 & 0.911 \\
GPT-4o-mini            & 0.065          & 0.896 & 0.614 \\
\bottomrule
\end{tabular}
\caption{Verifier comparison on 200 held-out pairs. Best in \textbf{bold}. Regression measures monotonicity violations; Parse measures valid JSON output rate; Agreement measures consistency with manual annotations.}
\label{tab:verifier_comparison}
\end{table}

\paragraph{Qualitative behavior.}
The accuracy gap between verifiers is largely driven by how they handle multi-step evidence integration. DeepSeek-v3.2 synthesizes evidence across distant search steps, while weaker verifiers tend to focus narrowly on the most recent observation and miss earlier critical evidence. The following example illustrates this pattern.



\begin{examplebox}{Verifier Judgment Comparison Example}
\textbf{Query:} \textit{A group that participated in constructing a major railway held an anti-discrimination rally at a government building. The rally related to a historic discriminatory law. This group has two major urban settlements in the country. Where did the final spike ceremony of that railway take place?}

\textbf{Knowledge Condition (C1):} \textit{The Canadian Chinese community participated in the construction of the Canadian Pacific Railway.}

\textbf{Analysis:} DeepSeek-v3.2 synthesizes evidence across Steps 1, 3, and 4, establishing a comprehensive causal chain linking laborers to the community. Gemini-2.5-Flash-Lite focuses narrowly on Step 2, missing earlier critical evidence. This pattern consistently contributes to DeepSeek-v3.2's lower regression rate and higher agreement with annotations.
\end{examplebox}
\begin{greenbox}{DeepSeek-v3.2 Judgment}
\textbf{DeepSeek-v3.2 Judgment (Status: 1):}
\begin{itemize}[leftmargin=*, itemsep=0.5pt, topsep=1pt, partopsep=2pt]
    \item \textbf{rationale:} The knowledge requires 'Canadian Chinese Community participated in the construction of Canadian Pacific Railway'. Evidence confirms: (1) Chinese laborers participated in building the CPR; (2) The community organizing the rally is the Canadian Chinese Community. Webpages explicitly state Chinese laborers built the CPR, and the rally was organized by Canadian Chinese community organizations. Chinese laborers are a subset of the Chinese community, and their construction participation is attributed to the community.
    \item \textbf{Evidence Chain:} [Step 1-Webpage 47] "In the late 19th century, a large number of Chinese laborers participated in the construction of the Canadian Pacific Railway" $\rightarrow$ [Step 3-Webpage 77] $\rightarrow$ [Step 4-Webpage 120] "Approximately 17,000 Chinese laborers participated in the construction of the Canadian Pacific Railway"
\end{itemize}
\end{greenbox}
\begin{redbox}{Gemini-2.5-Flash-Lite Judgment}
\textbf{Gemini-2.5-Flash-Lite Judgment (Status: 1):}
\begin{itemize}[leftmargin=*, itemsep=0.5pt, topsep=1pt, partopsep=0pt]
     \item \textbf{rationale:} Search results confirm Chinese laborers participated. Webpage 30: 'On November 7, 1885, the last spike was driven into the railway tie'. Webpage 32: 'Without Chinese laborers, the Pacific Railway could not have been completed on schedule'.
     \item \textbf{Evidence Chain:} [Step 2-Webpage 30] $\rightarrow$ [Step 2-Webpage 32]
\end{itemize}
\end{redbox}

\paragraph{Verification Prompt Template.}
The judge prompt used by all five verifiers is shown below.

\begin{examplebox}{Judge Prompt for Knowledge State Verification}
You are an expert evaluator assessing the multi-step information retrieval and reasoning progress of an AI Assistant.

You will be provided with:

1. [Question]: The original complex question.

2. [Knowledges]: A list of factual criteria (K1, K2, etc.). Each Knowledge is constructed as strict Entity-Relation-Entity statements (e.g., [Subject Entity] + [Action/Relation] + [Object Entity/Location]).

3. [Cumulative Context]: The Assistant's historical trajectory, including search queries, retrieved web documents (\textless tool\_response\textgreater), and internal reasoning (\textless think\textgreater).

Your task is to act as a PEDANTIC and ULTRA-STRICT judge. Determine the satisfaction status of EACH Knowledge based SOLELY on the [Cumulative Context].

\noindent\textbf{\#\#\# CRITICAL EVALUATION RULES:}

\textbf{1. Entity-Level Granularity (Zero Semantic Drift):} You must strictly distinguish between different entity scopes and types. Do NOT substitute entities with loose semantic approximations.

- Example: "Chinese Community" is an organization/collective entity. It is strictly NOT equal to "Chinese laborers". If the knowledge requires "Community" but the text only mentions "laborers", it is a MISMATCH (Status 0).

\textbf{2. Strict "AND" Logic for Triples:} Break down the knowledge into sub-constraints (Subject, Action/Relation, Object/Location, Time). If even a single sub-constraint or entity is missing or generalized, the entire knowledge MUST be marked as 0.

\textbf{3. Constrained Jigsaw Rule (No Forced Associations):} You can combine evidence across different steps or webpages, BUT there MUST be an explicit logical bridge in the text. You CANNOT stitch together disconnected facts based on your own assumptions.

\textbf{4. Validating \textless think\textgreater\ (Parametric Knowledge):} The Assistant's \textless think\textgreater\ block is a valid source of truth, provided the statement is a definitive, affirmative factual claim. Guesses or questions in \textless think\textgreater\ are invalid.

\textbf{5. Zero Inference \& Anti-Hallucination:} Do NOT use your own external knowledge to fill in blanks. If you cannot quote the exact text proving the specific entity and relation, mark it as 0.

\textbf{6. Minimal Sufficient Evidence Chain (No Redundancy):} When a knowledge is marked as 1, you MUST extract the exact quotes that form the complete logical chain. DO NOT output multiple redundant quotes that prove the same point from different webpages. Only extract the MINIMAL but COMPLETE text segments necessary to verify all constraints in the knowledge, showing the logical flow.

\textbf{7. Evidence-Action Alignment (Action-Evidence Alignment):} This is the most crucial point. If a knowledge is rated as 1, its evidence must be derived from the current or historical \textless tool\_response\textgreater. If the fact exists only in the Assistant's \textless think\textgreater\ and lacks external retrieval evidence support, it must be rated as 0, unless the fact is common knowledge. This is to prevent the model from bypassing the search step by using parameter knowledge (Shortcut).

\noindent\textbf{\#\#\# OUTPUT FORMAT (JSON ONLY)}

You must output a strictly valid JSON object. Do not include markdown formatting like ```json. The JSON must follow this exact structure:

\{

\ \ "evaluations": \{

\ \ \ \ "K1": \{

\ \ \ \ \ \ "status": 0 or 1,

\ \ \ \ \ \ "rationale": "Verify constraints step-by-step. Example: 'The text mentions Chinese laborers , which is a mismatch for the required entity Chinese Community. Due to entity mismatch, status is 0.'",

\ \ \ \ \ \"evidence\_chain": "Provide the minimal, non-redundant exact quotes forming the complete logical chain (e.g., '[Step 1 - Webpage 2] quote A -> [Step 3 - \textless think\textgreater] quote B'). Leave empty "" if status is 0."

\ \ \ \ \},

\ \ \ \ "K2": \{

\ \ \ \ \ \ "status": 0 or 1,

\ \ \ \ \ \ "rationale": "...",

\ \ \ \ \ \ "evidence\_chain": "..."

\ \ \ \ \}

\ \ \}

\}

\noindent\textbf{[Question]} \{question\}

\noindent\textbf{[Knowledges]} \{knowledges\}

\noindent\textbf{[Cumulative Context]} \{cumulative\_text\}

\end{examplebox}

\section{Grouping Quality: Extended Results}
\label{app:gigpo-limitations}

This appendix expands the analysis of Section~\ref{sec:grouping_anchor}. We formally define the four grouping-quality metrics used in Table~\ref{tab:diag-main} and present a representative case where observation-based and knowledge-based grouping disagree.

\paragraph{Setup recap.}
We sample 500 queries from Xbench and BrowseComp, rollout the SFT checkpoint up to 16 times per query, and obtain 5,922 valid trajectories and 55,571 steps after filtering. Three anchors are compared: \emph{observation-anchored}, using GiGPO's original protocol (BGE-large cosine similarity over retrieved contents, threshold $\tau=0.9$); \emph{action-anchored}, using the same similarity protocol over generated queries; and \emph{knowledge-anchored} (ours), requiring exact equality of accumulated knowledge state.

\paragraph{Metric definitions.}
Let $\mathcal{G}$ denote the set of all groups produced by an anchor. We report:
\begin{itemize}[leftmargin=*, itemsep=2pt, topsep=2pt]
\item \textbf{Effective step ratio (ESR):} the fraction of steps in non-singleton groups,
\begin{equation}
\mathrm{ESR} = \frac{\sum_{g \in \mathcal{G}} |g| \cdot \mathbb{I}(|g| \geq 2)}{\sum_{g \in \mathcal{G}} |g|}.
\end{equation}
\item \textbf{Singleton ratio (SR):} the fraction of groups containing only one step,
\begin{equation}
\mathrm{SR} = \frac{|\{g \in \mathcal{G} : |g| = 1\}|}{|\mathcal{G}|}.
\end{equation}
\item \textbf{Mean non-singleton size (MeanNG):} the average size of non-singleton groups,
\begin{equation}
\mathrm{MeanNG} = \frac{\sum_{g \in \mathcal{G},\, |g| \geq 2} |g|}{|\{g \in \mathcal{G} : |g| \geq 2\}|}.
\end{equation}
\item \textbf{Cross-rollout coverage (Cross):} the average number of distinct rollouts contributing to each non-singleton group. This metric distinguishes genuine cross-trajectory alignment from local within-trajectory smoothing.
\item \textbf{$n_{\mathrm{groups}}$:} The size of the sample group associated with each grouping anchor.
\end{itemize}
\paragraph{Qualitative case.}
The following example illustrates why surface similarity is unreliable. Trajectory 6 has acquired 0 of 12 knowledge conditions (still in exploration), while Trajectory 10 has acquired 8 of 12 (the target entity has been identified). Yet the retrieved-content embeddings of the two trajectories' steps reach cosine similarity 0.987 because both happen to mention similar generic terms (Chinese fir, forest research, research laboratory). Observation-anchored grouping clusters the two steps together and ends up comparing an exploration step with a verification step. Knowledge-anchored grouping correctly separates them because their knowledge states differ.

\begin{examplebox}{Example }

\textbf{Question:} \textit{A forester from a specific country studied under a scholar who graduated from a now-defunct university in a certain city. This mentor served as the dean of agriculture at a regional national university, authored a monograph on a local island's forestry, and passed away in the mid-20th century. The forester dedicated his career to researching a specific plant, established a national research lab for it, and proposed an ecological theory. He also collaborated with a researcher who advocated for forest road construction and led a plant introduction survey. The forester eventually received an international award. In what year was this forester born?}

\textbf{Ground Truth:} 1915

\textbf{Knowledge conditions:} 
\begin{itemize}[leftmargin=2.5em, nosep, itemsep=1.5pt]
    \item[\textbf{\small\color{gray}K1.}] Xiong Wenyu is from China in East Asia and studied under the scholar Cheng Fuxin.
    \item[\textbf{\small\color{gray}K2.}] Cheng Fuxin graduated from Yenching University, a Christian university in the capital city that was later dissolved and merged.
    \item[\textbf{\small\color{gray}K3.}] Cheng Fuxin served as the dean of the College of Agriculture at National Sichuan University, a national university in Southwestern China.
    \item[\textbf{\small\color{gray}K4.}] Cheng Fuxin authored a monograph on the forestry of Taiwan, China's largest southeastern island.
    \item[\textbf{\small\color{gray}K5.}] Cheng Fuxin passed away in the mid-1950s.
    \item[\textbf{\small\color{gray}K6.}] Xiong Wenyu dedicated his career to researching bamboo, a plant described as humble and principled.
    \item[\textbf{\small\color{gray}K7.}] Xiong Wenyu established China's first bamboo research lab, the Bamboo Research Laboratory.
    \item[\textbf{\small\color{gray}K8.}] Xiong Wenyu proposed the ecological boundary layer theory, the Ecological Boundary Layer Theory.
    \item[\textbf{\small\color{gray}K9.}] Xiong Wenyu collaborated with researcher Zhou Fangchun, who advocated for forest road construction.
    \item[\textbf{\small\color{gray}K10.}] Zhou Fangchun led the bamboo introducing southern species to the north survey project, the Southern Bamboo Moving North Survey Project.
    \item[\textbf{\small\color{gray}K11.}] Xiong Wenyu received the International Bamboo Pioneer Award, the International Bamboo Pioneer Award.
    \item[\textbf{\small\color{gray}K12.}] Xiong Wenyu was born in 1915.
\end{itemize}

\textbf{Trajectory 6, Step 2: }

\indent \textbf{<think>} I need to match the eligible mentors and foresters more precisely. Let's review the key points and try a more targeted search... \textbf{</think>} 
\textbf{<tool\_call>} \{name: "search", arguments: \{query\_list: ["founder of Chinese fir research laboratory China", "founder of ginkgo research laboratory China", ...]\} \}\textbf{</tool\_call>}
\indent \textbf{<observation>} A search for "founder of the Chinese fir research laboratory" found the following results...\textbf{</observation>}

\noindent\textbf{Trajectory 10, Step 8: }

\indent \textbf{<think>} Based on the search results, there is no explicit mention of "recommending the construction of forest roads" among Xiong Wenyu's collaborators. Therefore, I need to search more information to judge...\textbf{</think>} 

\noindent \textbf{<tool\_call>} \{name: "search", arguments: \{query\_list: ["founder of Chinese fir research laboratory China", ...]\} \} \textbf{</tool\_call>}
\indent \textbf{<observation>} A search for "forester who recommended forest road construction" found the following results... \textbf{</observation>}
\end{examplebox}
\begin{greenbox}{Observation-Anchored Grouping}
\textbf{Tra6\_Step2\_state:} [0,0,0,0,0,0,0,0,0,0,0,0]

\textbf{Tra10\_Step8\_state:} [1,1,1,1,1,0,1,1,1,1,0,0]

\textbf{Conclusion:} Due to the discrepancy in their state vectors, Tra6\_Step2 and Tra10\_Step8 are not clustered into the same group.
\end{greenbox}

\begin{redbox}{Knowledge-Anchored Grouping}
\textbf{Similarity:} 0.9868339896202087.

\textbf{Conclusion:} With the similarity greater than the threshold, tra6\_step3 and tra10\_step8 are clustered into a single group.
\end{redbox}

\section{Implementation Details}
\label{app:implementation}

This appendix provides the full implementation details of KnowGPO and its baselines, organized into three subsections: tool environment, training configuration, and prompt formats.

\subsection{Tool Environment}
\label{app:implementation-tools}

The interactive environment exposes a single \texttt{search} action that routes queries to Xhub Search for Chinese queries or Google Search for English queries, with language detected via Unicode character ranges. The engine accepts batch inputs and returns the top-10 results per query, each comprising a title, URL, and snippet. To bound the context length consumed by retrieved evidence, snippets are compressed to an average of 300 tokens by a fine-tuned Qwen3-8B summarization model that preserves entities and key claims while removing boilerplate. Temporal sensitivity detection biases retrieval toward recent content for time-sensitive queries.

\subsection{Training Configuration}
\label{app:implementation-training}

\paragraph{Supervised fine-tuning.}
We first SFT Qwen3-8B on a corpus of approximately 8{,}100 deep search trajectories, constructed by combining human demonstrations with model-assisted annotation. SFT runs on 8 A100 GPUs using LLaMAFactory with standard causal language modeling, for 3 epochs at learning rate $2 \times 10^{-5}$ and effective batch size 64. The resulting checkpoint achieves over 98.5\% format accuracy on ReAct-style tool invocations and is the shared initialization for all RL methods we compare.

\paragraph{Reinforcement learning.}
Following SFT, we apply KnowGPO and the three RL baselines. Training data is selected from the same source distribution as SFT but filtered to the \textit{zone of proximal development}, defined as queries with pass@5 $\in [0.4, 0.6]$ on the SFT checkpoint. This yields 649 high-quality training queries. RL runs on 64 NVIDIA H20 GPUs using verl at learning rate $1 \times 10^{-6}$, batch size 32 prompts with 16 rollouts each, KL penalty $\beta = 0.01$, for 3 training epochs. All RL methods share these settings; they differ only in the advantage estimator and reward design.

\subsection{Prompt Formats}
\label{app:implementation-prompts}

The system prompt defines the available tools and tool-calling format. The user prompt enforces language consistency and structured answer formatting.

\begin{examplebox}{System Prompt for Training and Inference}
You are a helpful and harmless assistant.

\# Tools

You may call one or more functions to assist with the user query.

You are provided with function signatures within \textless tools\textgreater\textless/tools\textgreater\ XML tags:

\textless tools\textgreater

\{"type": "function", "function": \{"name": "search", 

\ \ "description": "Search the web for information. The result will be a string.", 

\ \ "parameters": \{"type": "object", 

\ \ \ \ "properties": \{"query\_list": \{"type": "array", 

\ \ \ \ \ \ "description": "Generate up to 5 search queries based on the requirements below:

\ \ \ \ \ \ \ \ - All queries should use generalized, search engine-friendly keywords.

\ \ \ \ \ \ \ \ - Avoid relative terms (e.g., 'third author'); use universal terms (e.g., 'author list').

\ \ \ \ \ \ \ \ - Keep queries short ($\leq$15 chars). Remove auxiliary words.

\ \ \ \ \ \ \ \ - For non-English sources, search in the original language.

\ \ \ \ \ \ \ \ - Search 1-5 queries in parallel. Split multi-entity questions.

\ \ \ \ \ \ \ \ - Do not re-search previous queries.", 

\ \ \ \ \ \ "item": \{"type": "string"\}

\ \ \ \ \}\}, 

\ \ \ \ "required": ["query\_list"]

\ \ \}

\}\}\}

\textless/tools\textgreater

For each function call, return a json object with function name and arguments within \textless tool\_call\textgreater\textless/tool\_call\textgreater\ XML tags:

\textless tool\_call\textgreater

\{"name": \textless function-name\textgreater, "arguments": \textless args-json-object\textgreater\}

\textless/tool\_call\textgreater
\end{examplebox}

\begin{examplebox}{User Prompt Template}
Your primary task is to solve the user's questions, leveraging the appropriate tools as needed.

\noindent\textbf{[Core Instruction: Language Consistency]}

You MUST write your entire response in the same language as the user's question.

- If the user asks in Chinese, you must think and write in Chinese.

- If the user asks in French, you must think and write in French.

- Do not switch to English unless the user's question was in English.

This applies to your reasoning, explanations, and the final answer.

\noindent\textbf{[Core Instruction: Answer Formatting]}

You MUST format your final answer using the following tags:

\textless answer\textgreater

[Your answer content here]

\textless/answer\textgreater

\noindent\textbf{[Core Instruction: Citation Formatting]}

If the search tool was used and you are referencing a specific webpage in your final answer (between the \textless answer\textgreater\ tags), you must use the format "[citation:x]", where x is the corresponding webpage number.

You must first engage in step-by-step, rigorous thinking and reasoning before using the appropriate tool or providing the final answer.

\noindent\textbf{\# The user's message is:} \{QUERY\}

\end{examplebox}

\section{Referenced Deep Search Agents}
\label{app:referenced-agents}

The scores reported for external deep search agents in Tables~\ref{tab:results_part1} and~\ref{tab:results_part2} are taken from their official technical reports or earlier publications: MiniMax-M2.1~\citep{minimax2026m21}, DeepSeek-V3.1/V3.2/R1~\citep{deepseek2025v31, deepseekai2025deepseekv32pushingfrontieropen,Guo_2025}, GLM-4.5/4.7~\citep{5team2025glm45agenticreasoningcoding, glm5team2026glm5vibecodingagentic}, Kimi-K2.5~\citep{kimiteam2026kimik25visualagentic}, Claude-4-Sonnet~\citep{claudesonnet45}, OpenAI o3~\citep{openai2025b}, WebExplorer-8B~\citep{liu2025webexplorerexploreevolvetraining}, WebSailor-7B/32B~\citep{li2025websailornavigatingsuperhumanreasoning}, ASearcher-Web-32B~\citep{gao2025turnsunlockinglonghorizonagentic}, DeepMiner-32B-RL~\citep{du2026openseekerdemocratizingfrontiersearch}, DeepDive-9B/32B~\citep{lu2025deepdive}, and MiroThinker-v1.0/v1.5/v1.7~\citep{miromindteam2026mirothinkerpushingperformanceboundaries, miromindteam2026mirothinker17h1heavyduty, meituanlongcatteam2026longcatflashthinking2601technicalreport}. We were unable to locate official references for GPT-5-High and Gemini-3.0-Pro at the time of writing and report numbers as released by their providers.

\section{Search Behavior: Case Study}
\label{app:search}

This appendix expands the search-behavior analysis in Section~\ref{sec:analysis} with the experimental setup of the search-turn statistics and a representative failure mode of GRPO that KnowGPO avoids.

\paragraph{Sampling.}
We randomly sample 500 test instances spanning BrowseComp-ZH, Xbench-2505, Xbench-2510, and GAIA, and partition them by joint solving outcome between KnowGPO and GRPO: KnowGPO-only (127), GRPO-only (89), both pass (103), and both fail (181). Average search turns per subset are reported in Table~\ref{tab:search_efficiency}.

\paragraph{Failure mode: over-search erases the answer.}
The case below is typical of the GRPO-only failure region and illustrates a behavior KnowGPO avoids. The query asks for a specific word quoted in an academic article. Both methods recover the correct word by step 8. KnowGPO terminates because its knowledge state is saturated; GRPO continues for four more turns, retrieves paraphrastic snippets containing competing surface tokens (\textit{soften, satirize}), and overwrites its earlier correct answer. The example shows that absent a stopping signal grounded in actual evidence, additional search can actively destroy a correct intermediate answer.

\providecommand{\trajtag}[2]{%
  \textcolor{#1}{\texttt{<#2>}}%
}
\providecommand{\trajtagend}[2]{%
  \textcolor{#1}{\texttt{</#2>}}%
}
\providecommand{\thinkblock}[1]{%
  \par\smallskip\noindent
  \trajtag{blue!55!black}{think}\,#1\,\trajtagend{blue!55!black}{think}\par
}
\providecommand{\actionblock}[1]{%
  \par\smallskip\noindent
  \trajtag{purple!65!black}{action}\,#1\,\trajtagend{purple!65!black}{action}\par
}
\providecommand{\obsblock}[1]{%
  \par\smallskip\noindent
  \trajtag{teal!60!black}{observation}\,#1\,\trajtagend{teal!60!black}{observation}\par
}
\providecommand{\elide}{\textcolor{gray!70!black}{\itshape\,\dots\,remaining content elided\,\dots\,}}

\begin{examplebox}{Over-Search Hurts: A Reasoning Reversal under Redundant Exploration}
\textbf{Query:} \textit{In Emily Midkiff's June 2014 article in a journal named for one of Hreidmar's sons that guarded his house, what word was quoted from two different authors in distaste for the nature of dragon depictions?}

\textbf{Gold Answer:} \textit{fluffy}


\textbf{Analysis:} Both methods found the gold answer \textit{fluffy} by step~8. KnowGPO correctly halted. Lacking a stopping signal, GRPO executed 4 redundant turns, retrieved misleading paraphrases (\textit{soften/satirize}), and \textbf{overwrote its correct answer} with \textit{softened}. Thus, GRPO's extra steps actively destroyed an already-formed correct answer.
\end{examplebox}

\begin{greenbox}{KnowGPO Trajectory (\textbf{Correct})}
\begin{itemize}[leftmargin=*, itemsep=4pt, topsep=2pt, partopsep=2pt]

    \item Step 1
        \thinkblock{Identify Emily Midkiff and her June~2014 article. The journal is named after one of Hreidmar's sons that guarded his house. In Norse myth those sons are Fafnir, Ot\'r and Regin --- so the journal is most likely Fafnir.}
        \actionblock{search(``Emily Midkiff June 2014 article'', ``Hreidmar's sons that guarded the house'')}
        \obsblock{Top hit is the U.\ of North Dakota faculty page for \emph{Emily Midkiff}, listing a 2014 article \textit{``Dragons are Tricksy: The Pragmatic Trickster and the Appearance of Dragons in Children's Literature''} in \textit{Fafnir: Nordic Journal of Science Fiction and Fantasy Research}, vol.~1, no.~2 (June 2014), pp.~35--47. ......}

    \item Step 2
        \thinkblock{The journal must therefore be Fafnir, the dragon-son of Hreidmar who guarded the family treasure; the article is ``Dragons are Tricksy''. Verify the issue and date before reading the body text.}
        \actionblock{search(``Emily Midkiff `Dragons are Tricksy' Fafnir 2014 issue PDF'')}
        \obsblock{Returns the official \textit{Fafnir} 1(2) 2014 issue page on \texttt{journal.finfar.org}, table of contents lists \textit{``Dragons are Tricksy''} on pp.~35--47, alongside an open-access PDF link.......}

    \item Step 3
        \thinkblock{Fetch the open-access PDF so the literal quoted word can be located in the article body.}
        \actionblock{fetch(``https://journal.finfar.org/.../dragons-are-tricksy.pdf'')}
        \obsblock{The full PDF loads. The relevant paragraph (p.~38) reads: \textit{``Both Stein (1968) and Blount (1974) complain about the way dragons in modern children's books have become, in their words, `fluffy' --- harmless, cuddly companions stripped of menace.''} ......}

    \item Step 4
        \thinkblock{The article quotes ``fluffy'' from two authors: \textit{Stein (1968)} and \textit{Blount (1974)}. Confirm the citation handles before treating this as the answer.}
        \actionblock{search(``Ruth Stein 1968 children's literature dragons'', ``Margaret Blount 1974 dragons'')}
        \obsblock{Stein hit: Ruth M.\ Stein, \textit{``The Changing Styles in Dragons --- from Fafnir to Smaug''}, \textit{Elementary English}, 45(2), 1968. Blount hit: Margaret Blount, \textit{Animal Land: The Creatures of Children's Fiction}, William Morrow, 1974. Both are bona fide primary sources cited by Midkiff. ......}

    \item Step 5
        \thinkblock{Cross-check Stein's 1968 paper to confirm she literally writes ``fluffy'', not a Midkiff paraphrase.}
        \actionblock{search(``Ruth Stein 1968 Elementary English dragon fluffy quote'')}
        \obsblock{JSTOR snippet of Stein 1968, p.~180: \textit{``\dots~the modern picture-book dragon is so `fluffy' that one can hardly believe it ever breathed fire.''} The word is in scare quotes in the original. ......}

    \item Step 6
        \thinkblock{Same cross-check for Blount's 1974 \textit{Animal Land}.}
        \actionblock{search(``Margaret Blount Animal Land 1974 fluffy dragon quotation'')}
        \obsblock{Google Books preview of \textit{Animal Land} (1974), p.~116: \textit{``\dots~the dragon has dwindled to a `fluffy' pet, robbed of all that once made him terrible.''} The literal scare-quoted word matches Stein's. ......}

    \item Step 7
        \thinkblock{Both primary sources independently use the literal word ``fluffy'' in distaste. The journal is \textit{Fafnir}, the article is \textit{``Dragons are Tricksy''}, and the two cited authors are Stein and Blount. All three knowledges in C\textsubscript{suf} are now verified.}
        \actionblock{verify(C\textsubscript{suf} = \{\textit{Fafnir}, Stein \& Blount, ``fluffy''\})}
        \obsblock{Internal sufficiency check returns \(f^{\text{cum}}=\mathbf{1}\); no remaining sub-claim is unsupported. ......}

    \item Step 8
        \thinkblock{Knowledge state is saturated --- further searching cannot improve the answer and risks introducing distractors. Stop searching.}
        \actionblock{answer(``fluffy'')}

\end{itemize}
\textcolor{green}{\Large \ding{51}}  \textbf{Outcome:} The answer is fluffy
\end{greenbox}

\begin{redbox}{GRPO Trajectory (\textbf{Incorrect})}
\begin{itemize}[leftmargin=*, itemsep=4pt, topsep=2pt, partopsep=2pt]

    \item Step 1
        \thinkblock{Identify Emily Midkiff's June~2014 article and the journal named after one of Hreidmar's sons that guarded his house}
        \actionblock{search(``Emily Midkiff June 2014 article'', ``Hreidmar's sons that guarded the house'')}
        \obsblock{Top hit: \emph{Emily Midkiff}, U.\ of North Dakota; 2014 paper \textit{``Dragons are Tricksy''} in \textit{Fafnir: Nordic Journal of Science Fiction and Fantasy Research}, 1(2), pp.~35--47......}

    \item Step 2
        \thinkblock{The journal is Fafnir, the dragon-son of Hreidmar; the article is ``Dragons are Tricksy''. Verify the issue.}
        \actionblock{search(``Emily Midkiff `Dragons are Tricksy' Fafnir 2014 issue PDF'')}
        \obsblock{Confirms \textit{Fafnir} 1(2), June 2014, pp.~35--47 on \texttt{journal.finfar.org}, with an open-access PDF link......}
        
    \item[]  
    \centerline{$\boldsymbol{\cdots\cdots}$}
    
    \item Step 8 
        \thinkblock{Both Ruth Stein and Margaret Blount use the word `fluffy' to express distaste for softened dragon depictions......}
        \actionblock{answer\_in\_step(``fluffy'')}
        \obsblock{\textit{Intermediate Answer:} \textit{``\dots the word quoted from two different authors in distaste for dragon depictions is \textcolor{red}{\textbf{fluffy}}\dots''} (\textcolor{red}{\emph{gold answer already produced here}})}

    \item Step 9
        \thinkblock{Continue searching to corroborate the answer with additional evidence.\,......}
        \actionblock{search(``Emily Midkiff Fafnir 1.2 PDF'', ``Midkiff dragons softened'', ``Midkiff dragons satirized'')}
        \obsblock{A new third-party blog post is retrieved: \textit{``\dots scholars have noted a trend to soften and satirize the dragon for children, turning a once-fearsome beast into a comic sidekick \dots''} --- a paraphrastic description, \emph{not a literal quotation}......}

    \item Step 10
        \thinkblock{The new tokens \textit{`soften'} / \textit{`satirize'} are prominent in this snippet --- maybe these are the actual quoted words after all.\,......}
        \actionblock{search(``Midkiff `softened' dragon quoted'')}
        \obsblock{Returns the same paraphrastic blog post recycled across pages, plus a few unrelated essays --- no literal author quotation supports \textit{``softened''}......}

    \item Step 11
        \thinkblock{The new snippets foreground the words \textit{`soften'} / \textit{`satirize'}; perhaps these are the quoted words after all.\,......}
        \actionblock{search(\dots ``softened dragons'', ``satirized dragons'')}
        \obsblock{More paraphrastic results; no source contains a literal scare-quoted ``softened'' attributed to either Stein or Blount......}

    \item Step 12
        \thinkblock{The quoted word should be `softened', used to describe the weakening of dragon imagery.\,......}
        \actionblock{final\_answer(``softened'')}

\end{itemize}
\textcolor{red}{\Large \ding{55}}  \textbf{Outcome:} The answer is soften.
\end{redbox}

\end{document}